# Developing Improved Greedy Crossover to Solve Symmetric Traveling Salesman Problem

**Hassan Ismkhan ( Esmkhan )[1], Kamran Zamanifar[2]**

**[1] Computer Department, University of Bonab ( Binab ), East Azerbaijan, Islamic Republic of Iran**

**[2] Computer Engineering Department, University of Isfahan, Isfahan, Islamic Republic of Iran**

**Abstract**

The Traveling Salesman Problem (TSP) is one of the most famous optimization problems. Greedy crossover designed by Greffenstette et al, can be used while Symmetric TSP (STSP) is resolved by Genetic Algorithm (GA). Researchers have proposed several versions of greedy crossover. Here we propose improved version of it. We compare our greedy crossover with some of recent crossovers, we use our greedy crossover and some recent crossovers in GA then compare crossovers on speed and accuracy.

## 1. Introduction

After introducing Genetic Algorithm (GA) by Holand [1] many GA crossover operators have been invented by researchers because the performance of GA depends on an ability of these operators. PMX [1] is one of first crossovers proposed by Goldberg and Lingle in 1985. Reference [14] stated some important shortcomings of PMX and to overcome them proposed extended PMX (EPMX). DPX [10] [11] is another crossover that produces child with greedy reconnect of common edges in two parents. Greedy Subtour Crossovers (GSXs) [7] [8] [9] family are another groups of crossovers that operate fast. GSX-2 [8] is improved version of GSX-0 [6] and GSX-1 [7].

In this paper we propose our Improved GX (IGX). We use IGX and some recent crossovers in GA to solve some TSPLIB's problems then we compare these crossovers on speed and accuracy. So rest of this paper organized as follows: In following section we represent some versions of GX. In section 3 we propose IGX. We represent GA in section 4. We put forward our experimental results in section 5 and summarize paper in 6.

## 2. Greedy Crossover (GX)

Major GXs select a node and copy it to child then it probes witch of its neighbors is nearest to it, so the nearest one is copied to child and this process is continued until child tour be completed. We show some previous versions of GX by example in Fig 2. In this example we use a graph with 8 nodes that its edges cost are as distance matrix in Fig 1.

| | 1 | 2 | 3 | 4 | 5 | 6 | 7 | 8 |
|---|---|---|---|---|---|---|---|---|
| 1 | 0 | 12 | 19 | 31 | 22 | 17 | 23 | 12 |
| 2 | 12 | 0 | 15 | 37 | 21 | 28 | 35 | 22 |
| 3 | 19 | 15 | 0 | 50 | 36 | 35 | 35 | 21 |
| 4 | 31 | 37 | 50 | 0 | 20 | 21 | 37 | 38 |
| 5 | 22 | 21 | 36 | 20 | 0 | 25 | 40 | 33 |
| 6 | 17 | 28 | 35 | 21 | 25 | 0 | 16 | 18 |
| 7 | 23 | 35 | 35 | 37 | 40 | 16 | 0 | 14 |
| 8 | 12 | 22 | 21 | 38 | 33 | 18 | 14 | 0 |

Fig. 1 Distance matrix

VGX operates more accurate than other versions but it is slow. Our purpose is to design new version of GX not only has more accurate but also operates fast.

## 3. Improved Greedy Crossover (IGX)

Previous versions of GX were slow or had not enough accuracy so we designed improved version of GX and named it improved greedy crossover (IGX). IGX is same to other versions of GX but in each step it probes only nodes that are not in child tour. To achieve this purpose we use two auxiliary double-linked list that each of them present one of parents tour. When a node is selected and copied to child tour it is eliminated from both double-liked lists. We show IGX in Fig.3. It can be easily seen that time complexity of IGX is O(n). It needs O(n) to form double-linked list and O(n) to complete child. Please consider that to complete child tour IGX need n steps and in each step it probes 4 nodes.

## 4. Genetic Algorithm (GA)

To compare crossovers we use each of them in GA to solve some of TSPLIB data set then we compare speed and accuracy of them. We define GA as below:

```
1) Initialize population with random tours
2) while  population is changed
3)        for i =1 to  Generation-Size
4)              Select father and mother from population
5)              child ← operate one of crossovers
6)              operate 2_Opt_move_based LS on child [14]
7)              operate 3_Opt_move_based LS on child [14]
8)              add child to population
9)        reduce population
10) return best individual of population
```

GA use random tours to initialize population in line 1. After that child is produced in line 5, it is improved by 2-opt and 3-opt and then it is added to population. Low quality tours are eliminated from population in line 9. "while" loop in line 2 repeated until no better child produced. If one of produced children is better than one of population individual then "while" loop will be continued.

| father: | 4 | 5 | 7 | 3 | 2 | 1 | 6 | 8 |
|---|---|---|---|---|---|---|---|---|

| mother: | 5 | 1 | 7 | 3 | 6 | 2 | 4 | 8 |
|---|---|---|---|---|---|---|---|---|

**step 1:**
First node is selected randomly and copied to tour.
References [3][4] always use same node at start.
please suppose that selected number would be 1.

| child: | 1 | | | | | | | |
|---|---|---|---|---|---|---|---|---|

**step 2:**

| father: | 4 | 5 | 7 | 3 | 2 | 1 | 6 | 8 |
|---|---|---|---|---|---|---|---|---|

| mother: | 5 | 1 | 7 | 3 | 6 | 2 | 4 | 8 |
|---|---|---|---|---|---|---|---|---|

In each step four neighbors of recent selected node are considered and which is closer to it is selected.
2, 6, 5 and 7 are neighbors of 1 and 2 is closer to it so is copied to child.

| child: | 1 | 2 | | | | | | |
|---|---|---|---|---|---|---|---|---|

Step 2 repeated until tour is being completed.

**Special cases**

| father: | 4 | 5 | 7 | 3 | 2 | 1 | 6 | 8 |
|---|---|---|---|---|---|---|---|---|

| mother: | 5 | 1 | 7 | 3 | 6 | 2 | 4 | 8 |
|---|---|---|---|---|---|---|---|---|

3, 1, 6 and 4 are neighbors of 2 and 1 are closer to it but 1 is already exist in child then we cannot copy it to child.

| child: | 1 | 2 | ? | | | | | |
|---|---|---|---|---|---|---|---|---|

In this cases:

- GX [2] selects next node randomly.
- GX [3][4][5][6] considers another three nodes.
- If all of four nodes would be in child then:
  - References [3][4] select next node randomly.
  - Reference [5] chooses closer node (to recent selected node) among 20 random remaining nodes that are not copied to child yet.
  - Reference [6] operates very greedy and selects closer node among all remaining nodes. this version has been named very greedy crossover (VGX).

Fig. 2 GXs review

## 5. Experiments and Results

We implemented all of algorithms with c# language and used .NET 2008 and ran all experiments on AMD Dual Core 2.6 GHZ. We used each of EPMX[14], GSX-2 [7], UHX[1] [15], VGX [6], DPX [10][11], PBX [12] and our IGX in GA to solve eil51, eil101, kroA100, kroA200, a280 and lin318 instances which are all from TSPLIB [17]. We ran GA with each of seven crossovers for thirty times. In all of these runs we set Population-Size=50 and Generation-Size=500. Table I shows our comparison results. In this table "Best length", "Average length" and "Worst length" show the best, average, and worst tour lengths respectively. "Number of repeat "while" loop in lines 2 to 9" column points out how many times lines 2 to 9 in Fig. 4 is executed also "Average Time" column gives the average running time in seconds. In "Best length", "Average length" and "Worst length" columns the values in parentheses is result of calculating

$$\frac{\text{cost of solution found} - \text{known optimum cost}}{\text{known optimum cost}} \times 100$$

[1] . This crossover is unnamed and operates heuristically so we name it Unnamed Heuristic crossover (UHX).

These results show that IGX has more accuracy than other six crossovers. Sixth column in table I shows that GSX-2 and EPMX have more average number of repeat “while” loop in lines 2 to 8 than other crossovers it means that they have high diversity and can produce many different tours also in attention to this column we can result that they are quick.

Fig.4 summarize fourth column of table 1 and show average length of tours obtained by GA when uses each of crossovers. Fig.5 outline last column of table 1 and show average time of GA when uses each of crossovers.

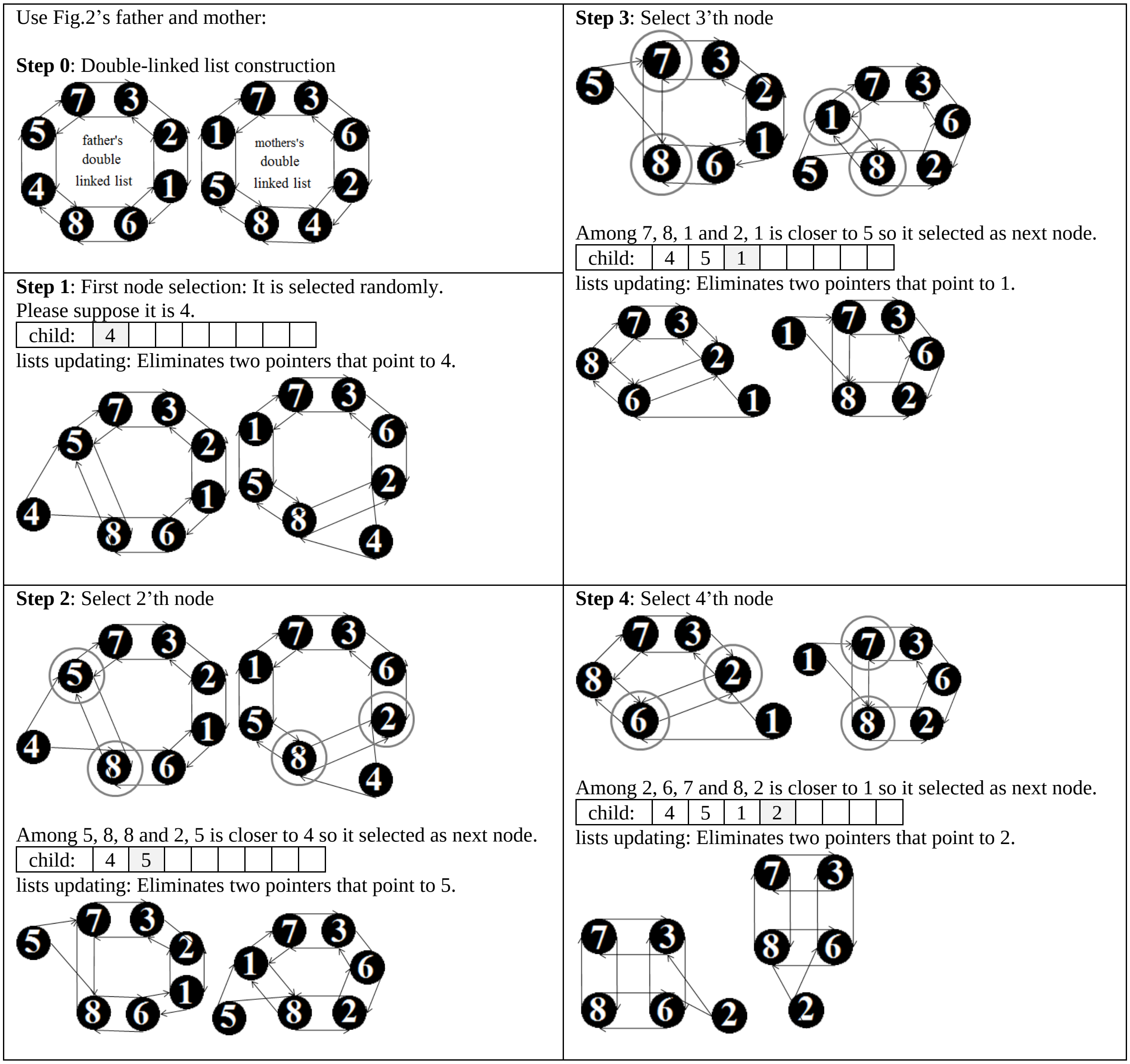

Figure 3: IGX

## 6. Conclusion

Greedy crossover (GX) designed by Greffenstette et al, is one of first heuristic crossover that can be used while Symmetric TSP (STSP) is resolved by Genetic Algorithm(GA). To improve its performance researchers have presented some versions of it but all of them are slow or has not enough accurate. In this paper we proposed new versions of it. IGX has more accuracy than not only any versions of GXs but any other considered crossover in this paper. In experiments we used IGX and six other recent crossovers in our GA to solve TSP instances. Experimental results have shown that when GA uses IGX has more accuracy than when uses other crossovers and also IGX is quick and complexity time of it is O(n).

Table 1 Experimental results

| Problem name | Crossover name | Best length (quality) | Average length (quality) | Worst length (quality) | Number of repeat “while” loop in lines 2 to 8 | Average Time(second) |
|---|---|---|---|---|---|---|
| Eil51 | EPMX | 433(1.64%) | 445(4.46%) | 459(7.75%) | **53** | 3.43044 |
| | GSX2 | 428(0.47%) | 446.1(4.72%) | 468(9.86%) | **42** | 1.56 |
| | UHX | 426(0%) | 430.5(1.06%) | 438(2.82%) | 29 | 5.31492 |
| | VGX | 430(0.94%) | 431.5(1.29%) | 434(1.88%) | 24 | 4.23228 |
| | DPX | 429(0.7%) | 431.5(1.29%) | 434(1.88%) | 21 | 1.51632 |
| | PBX | 429(0.7%) | 435.9(2.32%) | 445(4.46%) | 29 | 4.82352 |
| | IGX | 428(0.47%) | **428.8(0.66%)** | 431(1.17%) | 30 | 3.33216 |
| Eil101 | EPMX | 668(6.2%) | 684(8.74%) | 701(11.45%) | **110** | 10.7484 |
| | GSX2 | 671(6.68%) | 682.9(8.57%) | 698(10.97%) | **96** | 5.37888 |
| | UHX | 637(1.27%) | 649.7(3.29%) | 664(5.56%) | 43 | 15.1632 |
| | VGX | 631(0.32%) | 641.1(1.92%) | 653(3.82%) | 43 | 14.39256 |
| | DPX | 642(2.07%) | 653.3(3.86%) | 670(6.52%) | 29 | 4.77516 |
| | PBX | 670(6.52%) | 675.7(7.42%) | 687(9.22%) | 34 | 16.44864 |
| | IGX | 634(0.79%) | **640.5(1.83%)** | 652(3.66%) | 54 | 10.94496 |
| kroA100 | EPMX | 22295(4.76%) | 22959.6(7.88%) | 24013(12.83%) | **119** | 11.80764 |
| | GSX2 | 21940(3.09%) | 22492.5(5.69%) | 23068(8.39%) | **105** | 5.91396 |
| | UHX | 21320(0.18%) | **21440.4(0.74%)** | 21573(1.37%) | 42 | 15.21156 |
| | VGX | 21320(0.18%) | 21491.8(0.99%) | 21706(1.99%) | 45 | 15.08676 |
| | DPX | 21393(0.52%) | 21743.8(2.17%) | 23181(8.92%) | 34 | 5.031 |
| | PBX | 22603(6.21%) | 22915.3(7.67%) | 23392(9.91%) | 28 | 13.33332 |
| | IGX | 21292(0.05%) | 21510.7(1.07%) | 21794(2.41%) | 43 | 9.00588 |
| kroA200 | EPMX | 32347(10.14%) | 33264.8(13.27%) | 34297(16.78%) | **262** | 43.46472 |
| | GSX2 | 31378(6.84%) | 32437.8(10.45%) | 33440(13.87%) | **243** | 23.37192 |
| | UHX | 29680(1.06%) | 29950.6(1.98%) | 30872(5.12%) | 81 | 56.73096 |
| | VGX | 29706(1.15%) | 29995.6(2.14%) | 30392(3.49%) | 57 | 38.35104 |
| | DPX | 30079(2.42%) | 30532.2(3.96%) | 31077(5.82%) | 47 | 15.57192 |
| | PBX | 30996(5.54%) | 32285.7(9.93%) | 33679(14.68%) | 26 | 41.68788 |
| | IGX | 29649(0.96%) | **29773.3(1.38%)** | 29870(1.71%) | 43 | 17.03832 |
| A280 | EPMX | 2887(11.94%) | 3081(19.46%) | 3169(22.88%) | **380** | 79.69104 |
| | GSX2 | 2923(13.34%) | 3002.5(16.42%) | 3066(18.88%) | **364** | 44.57076 |
| | UHX | 2649(2.71%) | 2693.9(4.46%) | 2766(7.25%) | 80 | 62.65116 |
| | VGX | 2639(2.33%) | 2662.4(3.23%) | 2683(4.03%) | 69 | 61.1598 |
| | DPX | 2651(2.79%) | 2720.4(5.48%) | 2776(7.64%) | 49 | 24.33756 |
| | PBX | 2841(10.16%) | 2882.2(11.76%) | 2925(13.42%) | 40 | 115.60848 |
| | IGX | 2593(0.54%) | **2625.1(1.79%)** | 2654(2.91%) | 57 | 28.3842 |
| Lin318 | EPMX | 46956(11.72%) | 48373.9(15.1%) | 50058(19.1%) | **465** | 112.4214 |
| | GSX2 | 45971(9.38%) | 47307.6(12.56%) | 48573(15.57%) | **440** | 62.4078 |
| | UHX | 43354(3.15%) | 44003.7(4.7%) | 45078(7.25%) | 126 | 137.82756 |
| | VGX | 43293(3.01%) | 43756.6(4.11%) | 44327(5.47%) | 78 | 81.6816 |
| | DPX | 44381(5.6%) | 45052.9(7.19%) | 45814(9.01%) | 68 | 40.86732 |
| | PBX | 46940(11.68%) | 47843.6(13.83%) | 48479(15.35%) | 31 | 106.0176 |
| | IGX | 42992(2.29%) | **43486.1(3.47%)** | 44031(4.76%) | 75 | 44.93424 |

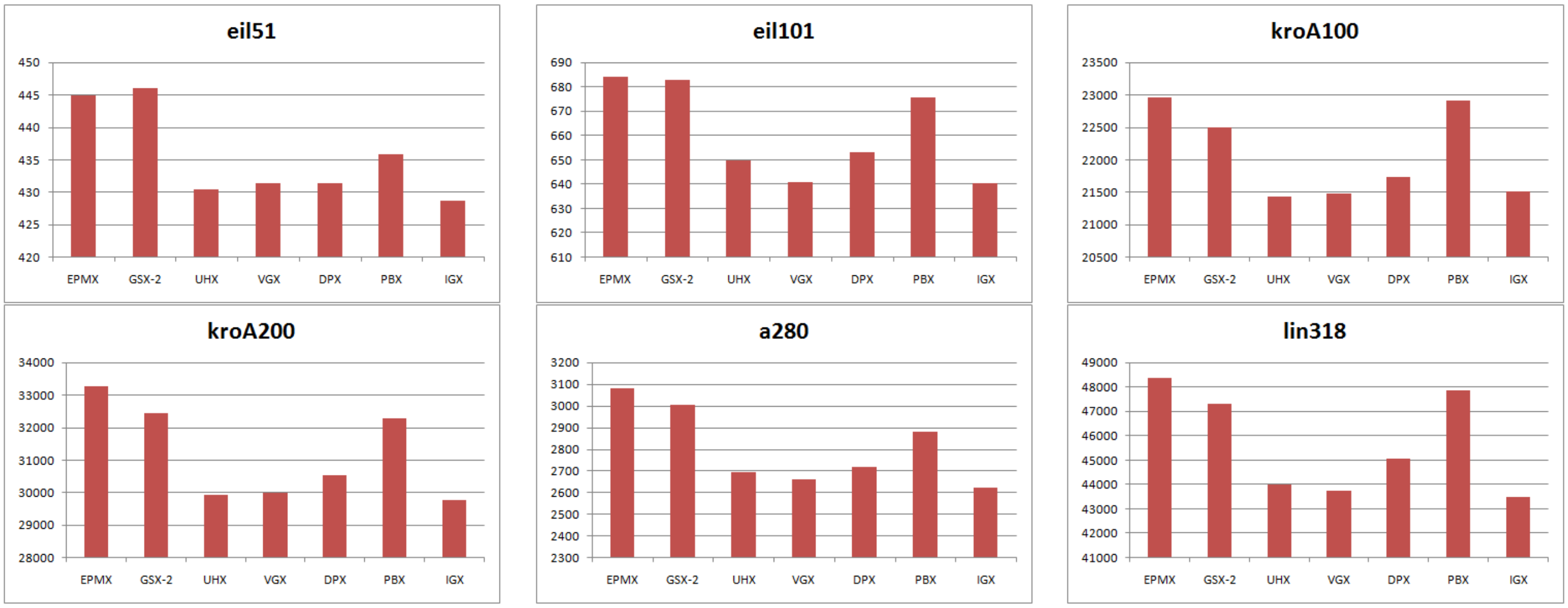

Figure 4. Average length

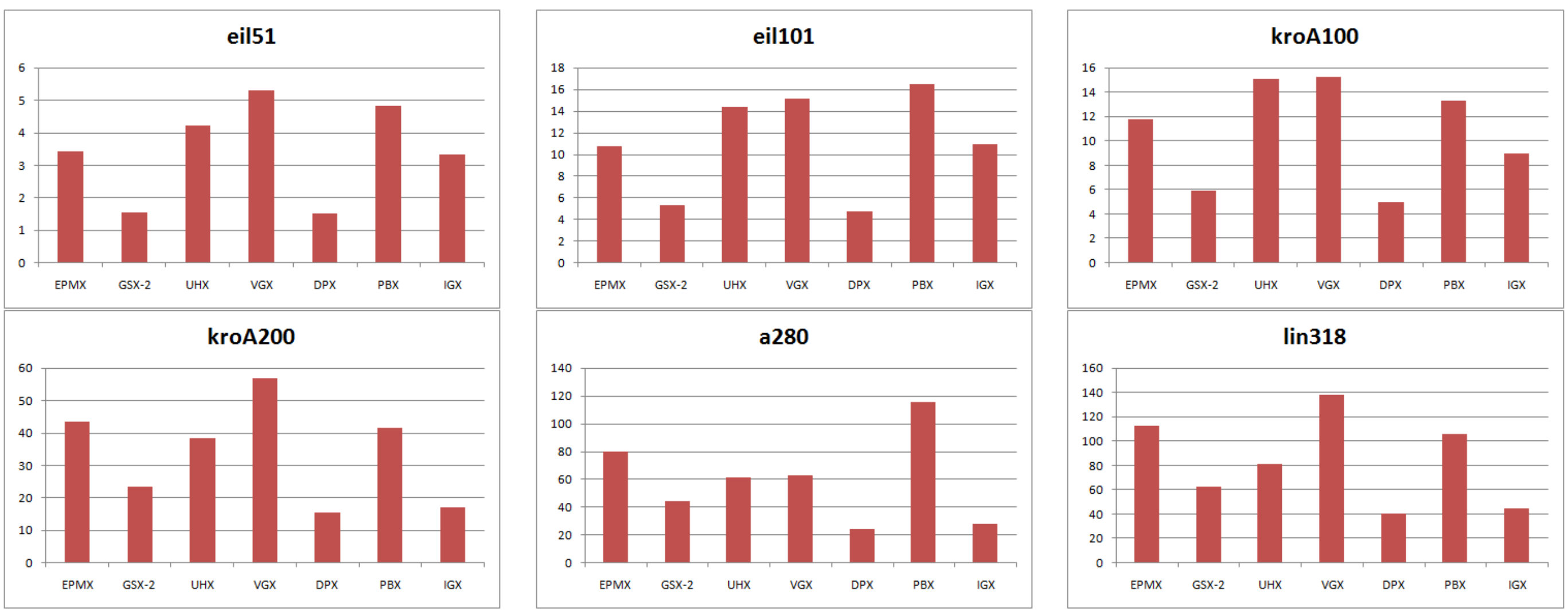

Figure 5. Average time of GA convergence when uses each of crossovers.

# References

[1] P.LARRANAGA, C.M.H.Kuijpers, R.H.Murga, I.Inza and S.Dizdarevic. "Genetic algorithms for the travelling salesman problem: A review of representations and Operators". Artificial Intelligence Review, vol.13, pp. 129-170, April.1999.

[2] J. J. Grefenstette, R. Gopal, B. Rosmaita, and D. VanGucht, "Genetic algorithms for the traveling salesman problem," in Proc. the 1st International Conference on Genetic Algorithms, 1985, pp. 160–168.

[3] G. E. Liepins, M. R. Hilliard, Mark Palmer and Michael Morrow, "Greedy genetics," in Proc. the Second International Conference on Genetic algorithms and their application, October 1987, pp.90-99.

[4] J.Y.Suh, D.V.Gucht, "Incorporating heuristic information into genetic search," in Proc. the Second International Conference on Genetic algorithms and their application, October 1987, pp.100-107.

[5] P.Jog, J.Y. Suh, and D.V.Gucht, "The effects of population size, heuristic crossover, and local improvement on a genetic algorithm for the traveling salesman problem," in Proc. the Third International Conference on Genetic Algorithms, 1989, pp.110-115.

[6] B.A.Julstrom. “Very greedy crossover in a genetic algorithm for the traveling salesman problem.” 1995 ACM symposium on applied computing, 1995, pp.324-328.

[7] H. D. Nguyen, I. Yoshihara, K. Yamamori, and M. Yasunaga, “Greedy genetic algorithms for symmetric and asymmetric TSPs,” IPSJ Trans. Math. Modeling and Appl., vol.43, no. SIG10 (TOM7), pp. 165–175, Nov. 2002.

[8] H. Sengoku and I. Yoshihara, “A fast TSP solver using GA on JAVA,” in Proc. 3rd Int. Symp. Artif. Life and Robot., 1998, pp. 283–288.

[9] A. Takeda, S. Yamada, K. Sugawara, I. Yoshihara, and K. Abe, “Optimization of delivery route in a city area using genetic algorithm,” in Proc. 4th Int. Symp. Artif. Life and Robot., 1999, pp. 496–499.

[10] B. Freisleben and P. Merz, “A Genetic Local Search Algorithm for Solving Symmetric and Asymmetric Traveling Salesman Problems,” in Proc. the 1996 IEEE International Conference on Evolutionary Computation, (Nagoya, Japan), pp.616-621, 1996.

[11] B. Freisleben and P. Merz, “New Genetic Local Search Operators for the Traveling Salesman Problem,” in Proc. the 4th Conference on Parallel Problem Solving from Nature – PPSN IV, (H.-M. Voigt, W. Ebeling, I. Rechenberg, and H.-P. Schwefel,eds.), pp. 890-900, Springer, 1996.

[12] H.Ismkhan and K.Zamanifar “Using Ants as a Genetic Crossover Operator in GLS to Solve STSP” SoCpar 2010, pp. 344-348.

[13] TSPLIB,http://www.iwr.uniheidelberg.de/groups/comopt/software/TSPLIB95/.

[14] Z.Tao. “TSP Problem solution based on improved Genetic Algorithm”. Fourth International Conference on Natural Computation. ICNC '08. Vol.1, pp.686-690, 2008.

[15] G.Vahdati, M.Yaghoubi, M.Poostchi and S.Naghibi. “A New Approach to Solve Traveling Salesman Problem Using Genetic Algorithm Based on Heuristic Crossover and Mutation Operator,” in Proc. SoCPaR, 2009, pp.112-116.